\documentclass[conference]{IEEEtran}
\IEEEoverridecommandlockouts
\usepackage{cite}
\usepackage{amsmath,amssymb,amsfonts}
\usepackage{graphicx}
\usepackage{textcomp}
\usepackage{xcolor}
\usepackage{multirow} 
\usepackage{float}  

\def\BibTeX{{\rm B\kern-.05em{\sc i\kern-.025em b}\kern-.08em
    T\kern-.1667em\lower.7ex\hbox{E}\kern-.125emX}}
\begin{document}

\title{Imitation Learning for Obstacle Avoidance Using End-to-End CNN-Based Sensor Fusion\\}

\author{\IEEEauthorblockN{1\textsuperscript{st} Lamiaa H. Zain}
\IEEEauthorblockA{
\textit{School of Engineering and} \\
\textit{Applied Science}  \\
\textit{Nile University}\\
Giza, Egypt \\
lzain@nu.edu.eg}
\and
\IEEEauthorblockN{2\textsuperscript{nd} Hossam H. Ammar}
\IEEEauthorblockA{\textit{School of Engineering and} \\
\textit{Computer Science} \\
\textit{University of Hertfordshire}\\
Administrative Capital, Egypt \\
h.ammar@gaf.edu.eg}
\and
\IEEEauthorblockN{3\textsuperscript{rd} Raafat E. Shalaby}
\IEEEauthorblockA{\textit{School of Engineering and} \\
\textit{Applied Science}\\
\textit{Nile University}\\
Giza, Egypt \\
rshalaby@nu.edu.eg}
}
\maketitle
\vspace{0.5em}
\noindent\textbf{Accepted Paper Notice:}  
This work has been accepted for presentation at the 2025 International Telecommunications Conference (ITC 2025). © 2025 IEEE. Personal use of this material is permitted. Permission from IEEE must be obtained for all other uses, including reprinting, republishing, or for creating new works.

\begin{abstract}
Obstacle avoidance is crucial for mobile robots' navigation in both known and unknown environments. This research designs, trains, and tests two custom Convolutional Neural Networks (CNNs), using color and depth images from a depth camera as inputs. Both networks adopt sensor fusion to produce an output: the mobile robot's angular velocity, which serves as the robot's steering command. A newly obtained visual dataset for navigation was collected in diverse environments with varying lighting conditions and dynamic obstacles. During data collection, a communication link was established over Wi-Fi between a remote server and the robot, using Robot Operating System (ROS) topics. Velocity commands were transmitted from the server to the robot, enabling synchronized recording of visual data and the corresponding steering commands. Various evaluation metrics, such as Mean Squared Error, Variance Score, and Feed-Forward time, provided a clear comparison between the two networks and clarified which one to use for the application.

\end{abstract}


\section{Introduction}
Mobile robot navigation is essential for various purposes, such as reaching specific destinations and avoiding obstacles. This navigation can be performed using either classic (conventional) methods or end-to-end approaches ~\cite{xiao2022motion}. Classic navigation employs sensory inputs like cameras, lidar, and sonar to create a representation of the environment (mapping). This map is then utilized by a global planner to determine local destinations, which are used by the local planner along with the environmental representation to produce the steering commands for mobile robots. Conversely, the end-to-end method generates low or high-level steering commands directly from sensory input data to reach a goal or maneuver around objects.

The classic approach involves multiple stages and considerable engineering effort. Consequently, the end-to-end method gains importance due to its ability to navigate at a higher level of semantic abstraction ~\cite{sepulveda2018deep} without needing to compute the robot's precise location or generate a map. Unlike the classic approach, which requires a predefined goal, the end-to-end method interacts with the environment in real-time. In this study, the end-to-end approach leverages two modalities, color and depth, with sensor fusion-based CNNs to directly predict steering commands and facilitate continuous motion.

The end-to-end strategy can be applied in three distinct ways:
\begin{itemize}
\item It can completely replace the conventional navigation stack. In this context, the entire navigation system functions as a black box, receiving sensory data and outputting steering commands.

\item It can also substitute intermediate navigation processes. For instance, spatiotemporal LSTMs can be used for environment perception \cite{li2022driver}, or CNNs can be employed for transfer learning to localize a humanoid robot \cite{9312592}. This approach has been reviewed in several surveys, including \cite{roy2021survey} and \cite{le2022survey}.

\item Finally, it can be incorporated as a phase in the enhancement process of the classic navigation stack \cite{cesar2021improving}.

\end{itemize} 

In this research, the third approach is adopted, where the end-to-end method is used for obstacle avoidance. The primary objective of this study is to train two different CNNs to predict the steering commands of a mobile robot. This is achieved using a real dataset obtained through a wireless communication setup between a remote server and the robot, with the robot navigating through different environments with different objects and varying lighting conditions. The mobile robot captured both color and depth images for training purposes. The wireless communication setup was essential for real-time control, synchronization of visual data, and accurate data labeling.

The dataset consists of 7,566 color images and 7,566 depth images, each of size 240x240 pixels, resulting in a total input size of approximately 1.745 GB. Training the networks on this dataset presented a challenge, but successful training with minimal errors was achieved. Additionally, the study aims to evaluate and compare the performance of the two CNNs on unseen datasets to determine the most suitable model for the task.

\section{Related Work}

This section provides a concise summary of similar research to this study.

One of the earliest studies applying the end-to-end approach was~\cite{thrun1995approach} that used an explanation-based neural network, EBNN, to navigate the Xavier mobile robot equipped with 24 sonar sensors, a laser light-stripper, and a color camera to reach a marked goal. Following Sebastian Thrun, many researchers followed the same path.

In the field of geometric navigation,~\cite{pfeiffer2017perception} and~\cite{9521224} replaced the total navigation stack by using CNNs. They had the goal as input to the model’s fully connected layer after the 2D laser range findings were fed to the network, and the outputs of the network were the robot’s linear and angular velocity.

For non-geometric navigation,~\cite{sepulveda2018deep} employed CNN as a supervised imitation learning application to move across various surroundings involving visual observations with subgoals as inputs and vehicle velocities as outputs. In another study,~\cite{zaccone2018random} modified the Rapidly-exploring Random Tree (RRT*)  to achieve path planning for a ship. 

For Hybrid Navigation,~\cite{zhou2019towards} combined the use of global and local planners for the navigation task, using a goal-directed supervised end-to-end approach where visual images and a one-hot encoded goal were used as input data to the network to generate high-level steering commands and a local planner that used deep reinforcement learning to avoid obstacles detected by a LIDAR.

Additionally, emphasizing the importance of sensor fusion, which combines inputs from multiple sensors to obtain the most accurate representation of the environment, ~\cite{housein2022extended} used an Extended Kalman Filter to merge odometry and depth data for localizing a TurtleBot. ~\cite{basavanna2021ros} integrated data from LIDAR and depth cameras to develop a more precise environmental map. Additionally, ~\cite{tran2021environment} employed sensor fusion of LIDAR and 3D ultrasonic sensors to create a 2D occupancy grid for mapping.

While numerous studies have utilized sonar sensors, depth sensors, LIDARs, and RGB cameras to collect environmental data for mobile robots, this work distinguishes itself by employing the Intel RealSense D415 depth camera, which combines both depth and color sensing in a single compact unit. This integration eliminates the need for multiple bulky sensors, offering a streamlined and efficient solution for real-time obstacle avoidance in both known and unknown environments.

The Intel RealSense D415, a high-performance model in the RealSense series, leverages stereo vision technology to deliver accurate depth measurements. Compared to traditional LIDAR systems, it offers significant advantages in cost, size, and ease of integration while still providing high-precision depth sensing. Unlike LIDAR, which typically requires a rotating mechanism to scan the environment, the D415 captures a wide field of view with real-time depth mapping through its stereo camera system, making it a practical alternative for mobile robot navigation.

Although LIDAR-based systems are known for their long-range precision, they generally require additional RGB cameras for semantic understanding, complicating sensor fusion workflows. These systems also introduce challenges in calibration, synchronization, and data alignment due to differing frame rates and coordinate systems. In contrast, the D415 delivers both RGB and depth data in a tightly coupled setup, simplifying the fusion process and reducing the computational overhead of aligning heterogeneous streams. While LIDAR remains advantageous for certain outdoor or long-range applications, the RealSense D415 provides a compact, cost-effective, and unified sensing solution particularly well-suited for indoor mobile robot navigation.

\section{Experimental Setup}

\subsection{Hardware Setup}

\begin{figure}[b]
  \centering
    \includegraphics[width=\linewidth]{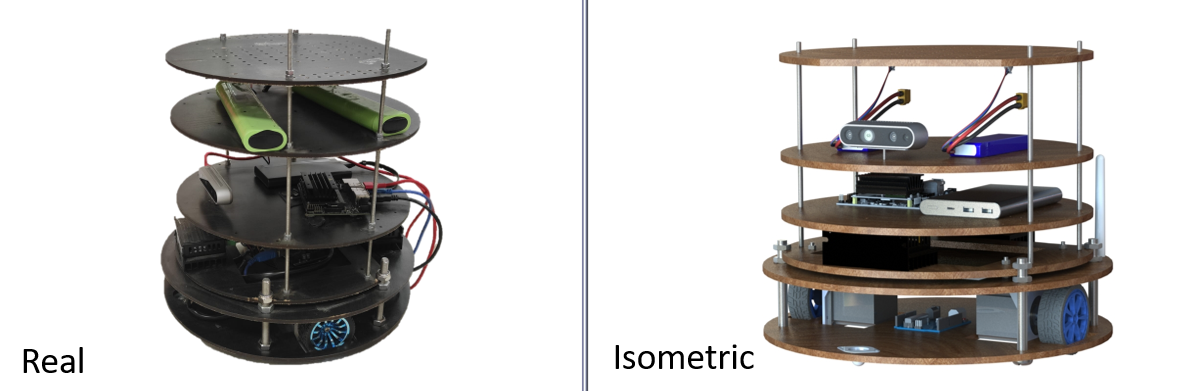}
    \caption{Mobile robot inspired by Turtlebot2.}
    \label{fig1}
\end{figure}

The used Mobile Robot was a differential drive mobile robot which had its design inspired by TurtleBot2 as shown in Fig.~\ref{fig1}. It has two wheels connected to two stepper motors and two metal caster wheels. The axis of the caster wheels is perpendicular to the axis connecting the two motors. The Mobile Robot is equipped with a Nividia Jetson Nano Developer Kit that acts as the on-robot computer and is used to deploy the CNN model in a real-time manner and generate angular velocities to the Arduino kit, which will, in sequence, send them to the Mobile Robot motors.
An IntelRealsense D415 Depth Camera was used for vision. It captures both color and depth data.

\subsection{Software Setup}
Ubuntu 18.04.6 was used as the operating system for both the Remote Server and the on-robot computer. Pytorch was used as a Python framework to train and test the designed architectures of the CNNs. ROS was used to facilitate communication between many different nodes like the Camera, the Arduino, and the Remote Server.

\begin{figure}[b]
  \centering
    \includegraphics[width=\linewidth]{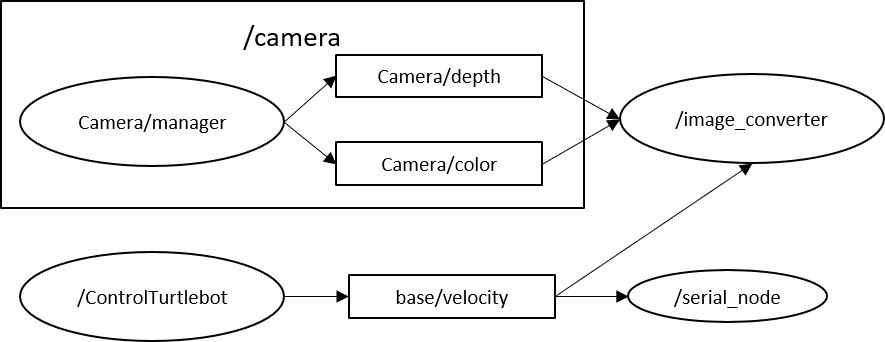}
    \caption{RQT ROS graph generated during data acquisition phase.}
    \label{fig2}
\end{figure}

\subsection{Robotic Operating System}

During the data collection phase, a human operator used the remote server to steer the mobile robot remotely using the ROS TurtleBot teleoperation package. The communication between the server and the on-robot computer is established using wireless communication protocols, WLAN. Fig.~\ref{fig2} shows the rqt\textunderscore ROS graph ~\cite{fairchild2016ros} that illustrates the active nodes and topics used. Names in ellipses are the nodes, and in rectangles are the topics. Color and depth topics, Camera/color and Camera/depth, are sent to an image converter node, /image\textunderscore converter,  to convert these topics' raw data to images. The image converter node saves these images alongside corresponding angular velocities subscribed from the /ControlTurtlebot node, launched by the remote server. Then, these velocities are sent to  Arduino and then to wheels via the Arduino serial node, /serial\textunderscore node.

\section{Methodology}

\subsection{Data Collection Stage}

During the process of data collection, the mobile robot was moving forward with a constant linear velocity of 0.1 m/s, and the angular velocity was controlled by a human operator. Angular velocities were allowed to change within the range of -0.3 to 0.3 rad/s with a step size of 0.1 rad/s. The total collected images were 7566 color images and 7566 depth images. Out of which, 60\% were used for training, 20\% for validation, and 20\% for testing. Mini batches of size 10 were used for training, 5 for validation, and 5 for testing. The whole dataset, including three RGB channels and a depth channel, was normalized to have a mean of 0 and a standard deviation of 1.

\begin{figure}[b]
  \centering
    \includegraphics[width=\linewidth]{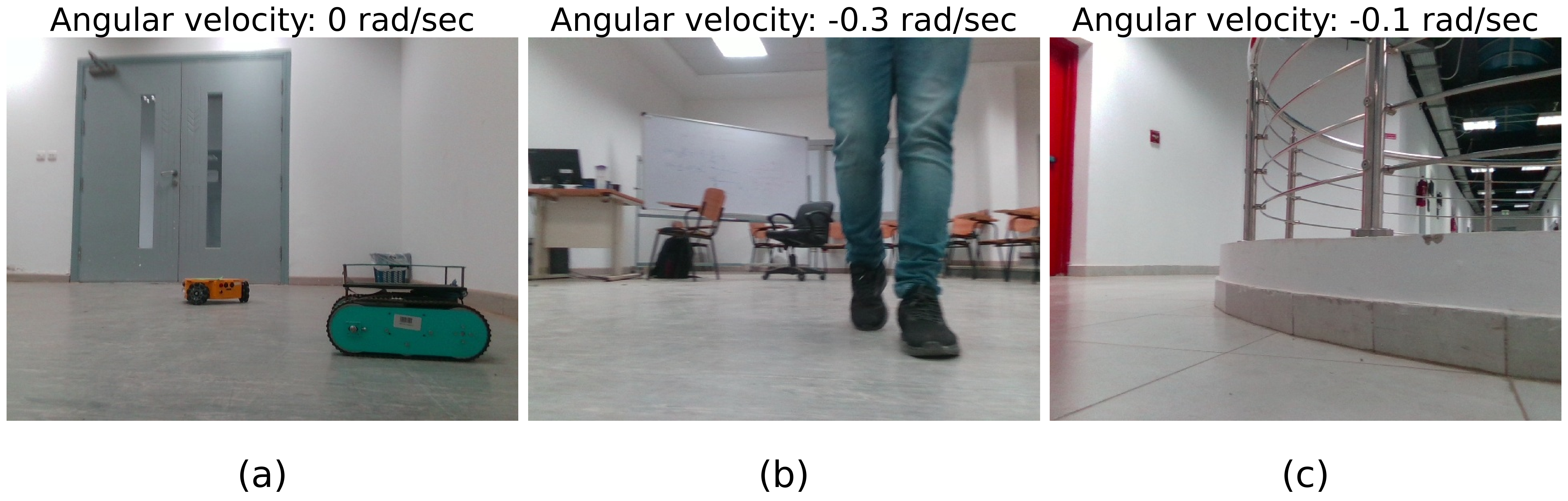}
    \caption{Sample images from different environments with various obstacles.}
    \label{fig3}
\end{figure}

Fig.~\ref{fig3} presents a sample of unnormalized images from the collected dataset, captured in diverse environments with varying lighting conditions and a combination of static and dynamic obstacles. In Fig.~\ref{fig3} (b), the dynamic obstacle is a human operator approaching the mobile robot.

\subsection{Proposed End-To-End Architecture}

Two networks were inspired by literature ~\cite{patel2019deep}. These networks are NetConEmb, and NetGated. The architecture of the two networks is re-illustrated in this research with a change in the input modalities where the depth data was collected via depth camera instead of a Lidar, which led to modifications to layers’ size. The RGB-D camera is configured to provide 240x240-pixel images.

Table~\ref{table1} briefly summarizes the difference between both architectures.

The architecture for both networks is a dual-branch Convolutional Neural Network (CNN) designed to process two different visual modalities: an RGB image and a depth image to leverage the complementary information from both inputs to improve perception and decision-making.

\begin{figure*}
  \centering
    \includegraphics[width=\linewidth]{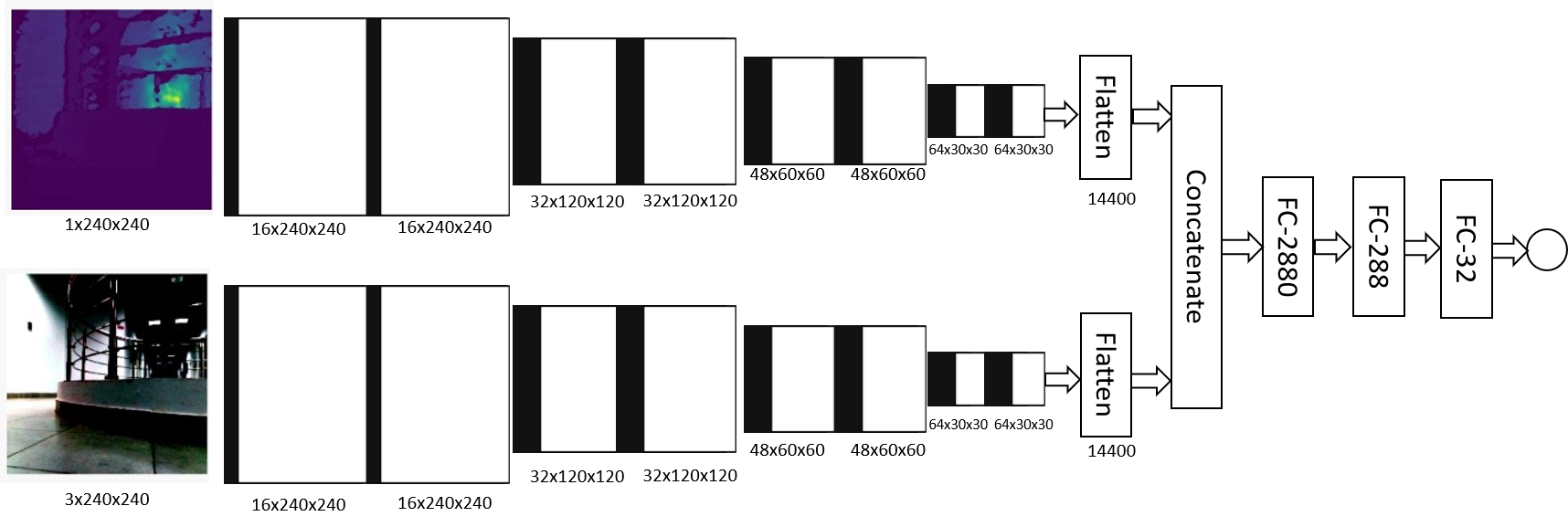}
    \caption{NetConEmb architecture.}
    \label{fig4}
\end{figure*}

\begin{figure*}
  \centering
    \includegraphics[width=\linewidth]{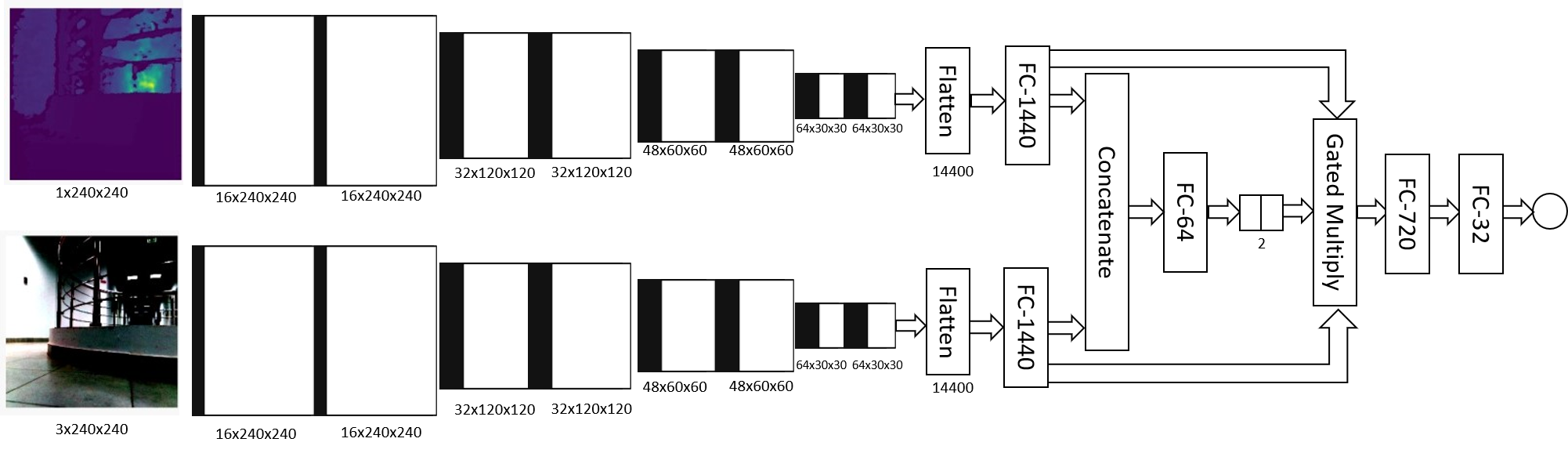}
    \caption{Netgated architecture.}
    \label{fig5}
\end{figure*}

 The architectures of both networks are shown in Fig.~\ref{fig4} and Fig.~\ref{fig5}, with the following common points:

 \begin{itemize}
        \item Top branch input is the depth image.
        \item Bottom branch input is the RGB image.
        \item Output of the final layer is the robot's angular velocity.
        \item For feature extraction, a series of Convolution layers and MaxPooling layers were used. 
        \item The convolution layers have a kernel or filter size of 3x3, a stride of 1, and padding is set to be the same to preserve the input dimensions.
        \item  A rectified linear unit (RELU) is used after each convolution layer to increase nonlinearity and to give the network the ability to learn and differentiate complex patterns in the dataset
        \item At the feature extraction stage, the depth and RGB branches are architecturally symmetric, enabling balanced feature learning from both modalities.
        \item The output feature maps from both CNN branches are flattened into one-dimensional feature vectors using the Flatten layer.
        
\end{itemize}

\begin{table}[b]
\caption{Comparison Between NetConEmb and NetGated Architectures}
\label{table1}
\begin{center}
\setlength{\tabcolsep}{6pt} 
\renewcommand{\arraystretch}{1.5} 
\begin{tabular}{|p{1.33cm}|p{2.8cm}|p{2.9cm}|}
\hline
\textbf{Aspect} & \textbf{NetConEmb} & \textbf{NetGated} \\

\hline

\textbf{Fusion Type} & {Simple Concatenation (early fusion)} & {Gated Fusion (adaptive, learned weighting)}\\
\hline

\textbf{Gating Mechanism} & {No gating at all – treats both modalities equally} & {Uses a Gated Multiply mechanism to modulate feature contributions from each modality}\\
\hline 

\textbf{Control logic over Modalities	} & {Fixed} & {Learnable – network decides how much attention to give each stream}\\
\hline   

\textbf{Fusion Complexity} & {Low} & {Higher (due to additional layers and gating logic)}\\
\hline  

\textbf{Architecture Complexity and width} & {Higher complexity (due to higher number of learned parameters) and wider} & {Low complexity and narrower}\\
\hline

\end{tabular}

\end{center}
\end{table}



Both Networks are different in the concatenation technique. In NetConEmb architecture, the flattened vectors from the depth and RGB streams are concatenated to form a single combined feature vector. The fused vector is passed through four fully connected (FC) layers:
\begin{itemize}
    \item FC-2880: A dense layer with 2880 output neurons.
    \item FC-288: A dense layer with 288 output neurons.
    \item FC-32: A dense layer with 32 output neurons.
     \item The final fully connected layer takes 32 neurons to produce the output angular velocity.
\end{itemize}

In NetGated, the flattened outputs are passed through the FC-1440 layers. The two 1440-dimensional vectors are concatenated. The concatenated vector is passed through the FC-64 layer to reduce dimensionality to a 64-dimentional vector, and then the following fully connected layer is used to produce the two learnable weights. Then, the gating mechanism is used. The gating mechanism has a Gated Multiply block that modulates the importance of each modality using the two learned weights by doing the following:

\begin{itemize}
    \item It multiplies the two learnable weights with their previous embedded feature vector.
    \item The outputs of the multiplication process are summed together before going through the FC-720 layer.
    \item This allows the network to "gate" the contribution of each stream before the final prediction.
    
\end{itemize}

The gated output is passed sequentially through a fully connected layer with 720 units (FC-720), followed by another with 32 units (FC-32), and finally through an output layer that predicts the angular velocity.

The gating mechanism of the NetGated architecture is described in Algorithm~\ref{alg:gating}, where scalar gating is employed to balance the contribution of each modality.
\begin{table}[b]
\caption{Gating Mechanism in NetGated Architecture}
\label{alg:gating}
\begin{center}
\begin{tabular}{|p{0.8\linewidth}|}
\hline
\textbf{Algorithm:} Gating Mechanism in NetGated Architecture \\
\hline
\textbf{Input:} $F_\text{rgb}$, $F_\text{depth}$ (Flattened feature vectors from RGB and Depth streams) \\
\hline
1. $E_\text{rgb} \gets \text{FC}_{1440}(F_\text{rgb})$ (Project RGB to 1440-d) \\
2. $E_\text{depth} \gets \text{FC}_{1440}(F_\text{depth})$ (Project Depth to 1440-d) \\
3. $F_\text{fused} \gets \text{Concatenate}(E_\text{rgb}, E_\text{depth})$ (Fused vector 2880-d) \\
4. $F_{64} \gets \text{FC}_{64}(F_\text{fused})$ (Reduce dimensionality to 64) \\
5. $[w_\text{rgb}, w_\text{depth}] \gets \text{FC}_{2}(F_{64})$ (Two scalar gating weights) \\
6. $E_\text{rgb-gated} \gets w_\text{rgb} \times E_\text{rgb}$ (Gated RGB features 1440-d) \\
7. $E_\text{depth-gated} \gets w_\text{depth} \times E_\text{depth}$ (Gated Depth features 1440-d) \\
8. $E_\text{gated} \gets E_\text{rgb-gated} + E_\text{depth-gated}$ (Summed gated vector 1440-d) \\
9. $F_\text{out} \gets \text{FC}_{720}(E_\text{gated})$ (Intermediate transformation 720-d) \\
10. $F_\text{out2} \gets \text{FC}_{32}(F_\text{out})$ (Intermediate transformation 32-d) \\
11. $\omega \gets \text{FC}_{1}(F_\text{out2})$ (Predicted angular velocity of the mobile robot 1-d) \\
\hline
\textbf{Output:} $\omega$ \\
\hline
\end{tabular}
\end{center}
\end{table}

\begin{figure}[t]
  \centering
    \includegraphics[width=\linewidth]{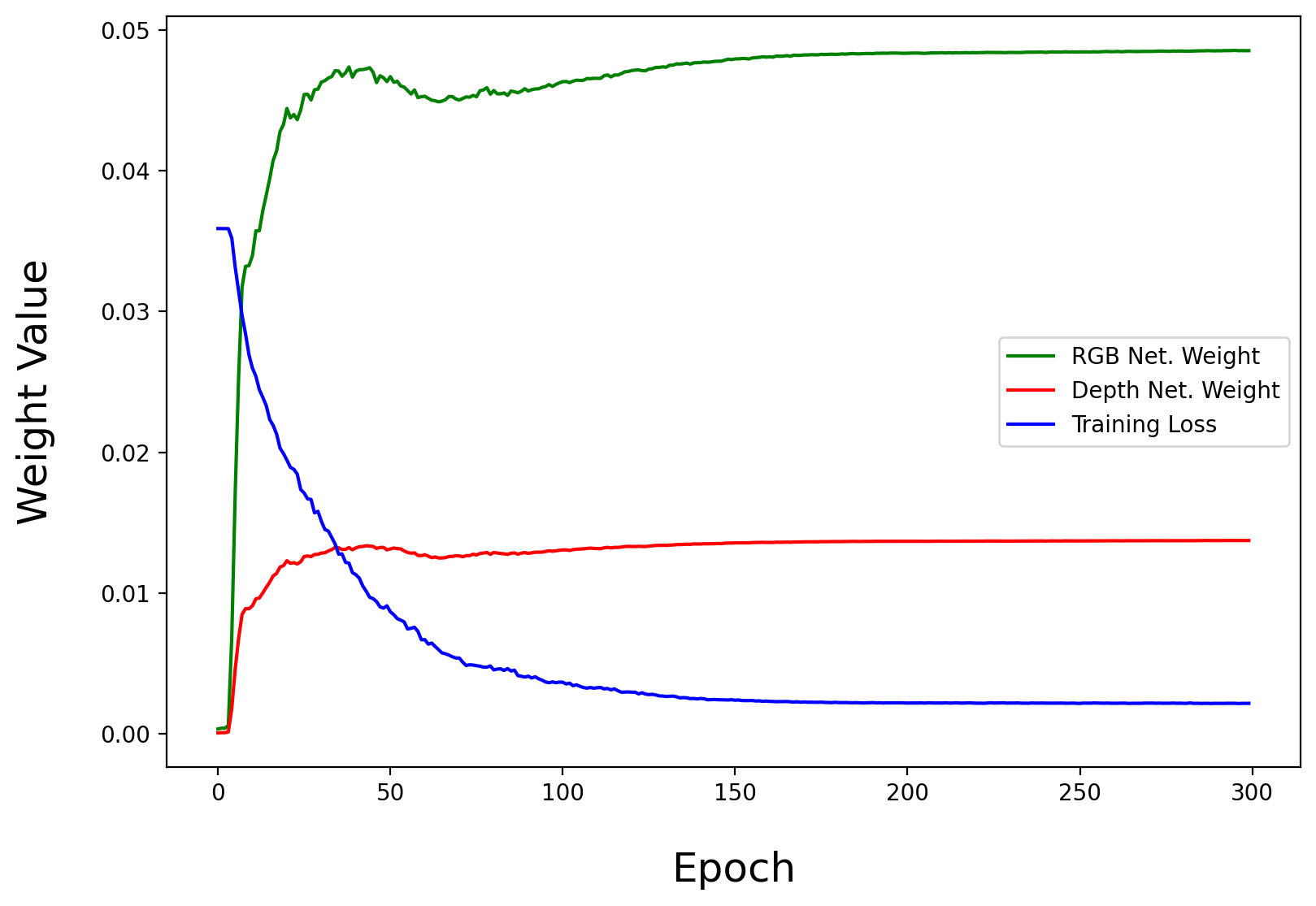}
    \caption{NetGated weights during the training process.}
    \label{fig6}
\end{figure}

Fig.~\ref{fig6} illustrates the evolution of the two scalar gating weights during the training process. The model learns optimal values for these weights based on the loss function. As the training loss decreases, both weights increase until convergence, at which point gradient updates become negligible. Throughout training, the RGB stream consistently receives a higher weight than the depth stream. Since the scalar weights in the NetGated model reflect modality importance, the higher RGB weight indicates that the model considers RGB features to be more informative and predictive of the target output (angular velocity). This observation aligns with expectations, as color images typically contain richer texture, color, and structural cues, particularly in well-lit environments. These results suggest that, had the model been trained in darker conditions (where RGB input is less reliable), the depth stream might have received a higher weight. This highlights the value of adopting a gating mechanism that dynamically adapts to the relative reliability of each modality across the dataset.

The total number of learned parameters in NetConEmb and NetGated was 83,902,001 and 22,098,691, respectively.

\subsection{Models' Training and Optimization Stage}

Since this problem is a regression problem, the criterion used in loss calculation was a regression criterion and was chosen to be Huber loss function \eqref{Eq:1} where ${a}$ refers to the residuals, i.e, the difference between actual and predicted values. ${\delta}$ was chosen to be 1 ${rad/s} $.

\begin{equation}
L_{\delta}(a)= \begin{cases}\frac{1}{2} a^{2} & \text { for }|a| \leq \delta \\ \delta\left(|a|-\frac{1}{2} \delta\right), & \text { otherwise }\end{cases} 
\label{Eq:1}
\end{equation} 

The adaptive moment estimation (ADAM) optimizer was used ~\cite{kingma2014adam}. ADAM combines the advantages of both Adaptive Gradient (AdaGrad) ~\cite{duchi2011adaptive} and Root Mean Square Propagation (RMSprop) suggested by Geoff Hinton, 2012 in one of his lectures.  ADAM shares with AdaGrad the advantage of changing the learning rate when sparse gradients are present. ADAM also shares with RMSprop the advantage of learning rate adaptivity with quickly changing gradients. 

\begin{figure}[t]
  \centering
    \includegraphics[width=\linewidth]{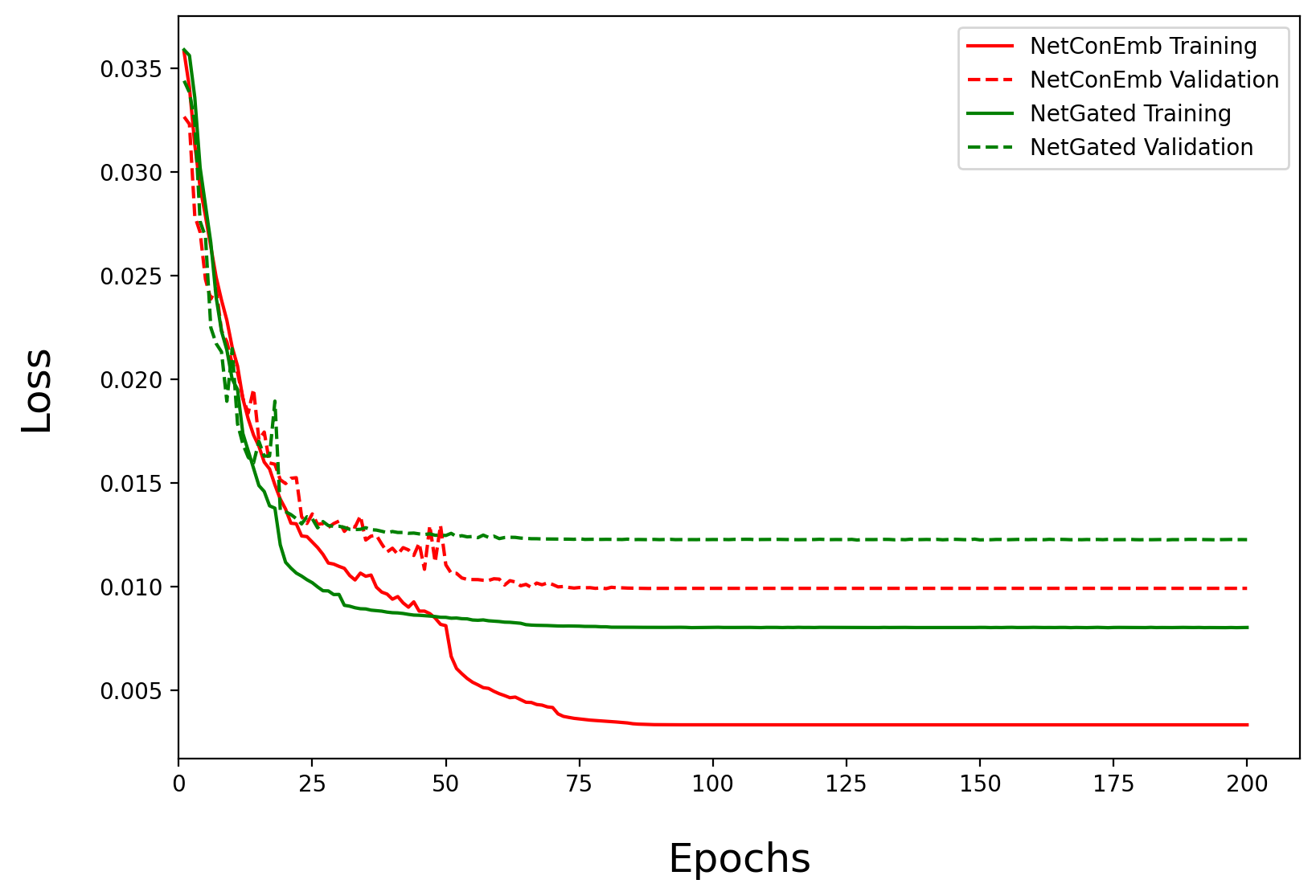}
    \caption{Training and validation losses for the first 200 epochs for NetConEmb and NetGated architectures.}
    \label{fig7}
\end{figure}

As illustrated in Fig.~\ref{fig7}, during the training process, loss kept decreasing successfully at each epoch where solid lines represent the training loss and dashed lines are validation loss. It's noticed that NetConEmb was able to reach lower training and validation losses than NetGated.
\begin{table}[b]
\caption{Different metrics of the trained networks’ behavior on the testing dataset}
\label{table2}
\begin{center}
\setlength{\tabcolsep}{6pt} 
\renewcommand{\arraystretch}{1.5} 
\begin{tabular}{|p{1.4cm}|p{0.9cm}|p{1cm}|p{1cm}|p{0.7cm}|p{1.1cm}|}
\hline
\textbf{Net/Metric} & \textbf{MAE $\times 10^3$} & \textbf{RMSE $\times 10^3$} & \textbf{MedAE $\times 10^3$} & \textbf{VS} & \textbf{Inference Time}\\
\hline

\hline
\textbf{NetConEmb} &  5.13 & 24.16 & 0.06 & 0.93 & 34.511 \\
\hline
\textbf{NetGated} &  6.11 & 25.49 & 0.08 & 0.92 & 41.518 \\
\hline
\end{tabular}
\end{center}
\end{table}


\section{Results}

In this section, the performance of each network is discussed based on testing each network against the testing dataset. Four evaluation metrics were used:

\begin{itemize}
    \item Mean Absolute Error (MAE): quantifies the overall error magnitude within the testing dataset.
    \item Root Mean Squared Error (RMSE): particularly sensitive to outliers and large errors.
    \item Median Absolute Error (MedAE): represents the median of all errors, providing robustness against outliers.
    \item Variance Score (VS): evaluates the model's ability to capture the dataset's variation, with a perfect score of 1.0 indicating optimal performance.
\end{itemize}

It is noteworthy that MAE, RMSE, and MedAE are measured in units of rad/s, whereas VS is dimensionless.

Also, the time for the feed-forward path of each network, inference time, was recorded in milliseconds. To determine the average duration, a batch consisting of five RGB images and five Depth images was used to generate five corresponding inferences. 

According to the findings presented in Table~\ref{table2}, NetConEmb architecture outperforms the NetGated architecture for this application. The NetConEmb architecture has less MAE, RMSE, and MedAE. Also, it has a higher variance score than the NetGated architecture. Also, considering the inference time, NetConEmb showed less inference time than the NetGated architecture.

Even though the NetConEmb is a more complex model with more learned parameters than the NetGated one, it had a shorter inference time. This needs to be explained. In NetConEmb, the architecture, which is a key factor that affects the inference time,  has simple sequential layers, which are highly optimized for parallel computation. However, the NetGated architecture contained a gating mechanism that uses custom operations (Gated Multiply and Summation), which requires more conditional logic and memory access and is less efficient on GPU/CPU during inference. Despite the fact that the gating mechanism in NetGated added extra intermediate tensors and more movement between layers, and that it contains more layers than the NetConEmb model after feature extraction, these layers are narrower; that is why they have fewer learned parameters, but they incur computational overhead which is more expensive per Floating Loint Operation (FLOP) and also contributed to the increase of the inference time.

\begin{figure}[b]
  \centering
    \includegraphics[width=\linewidth]{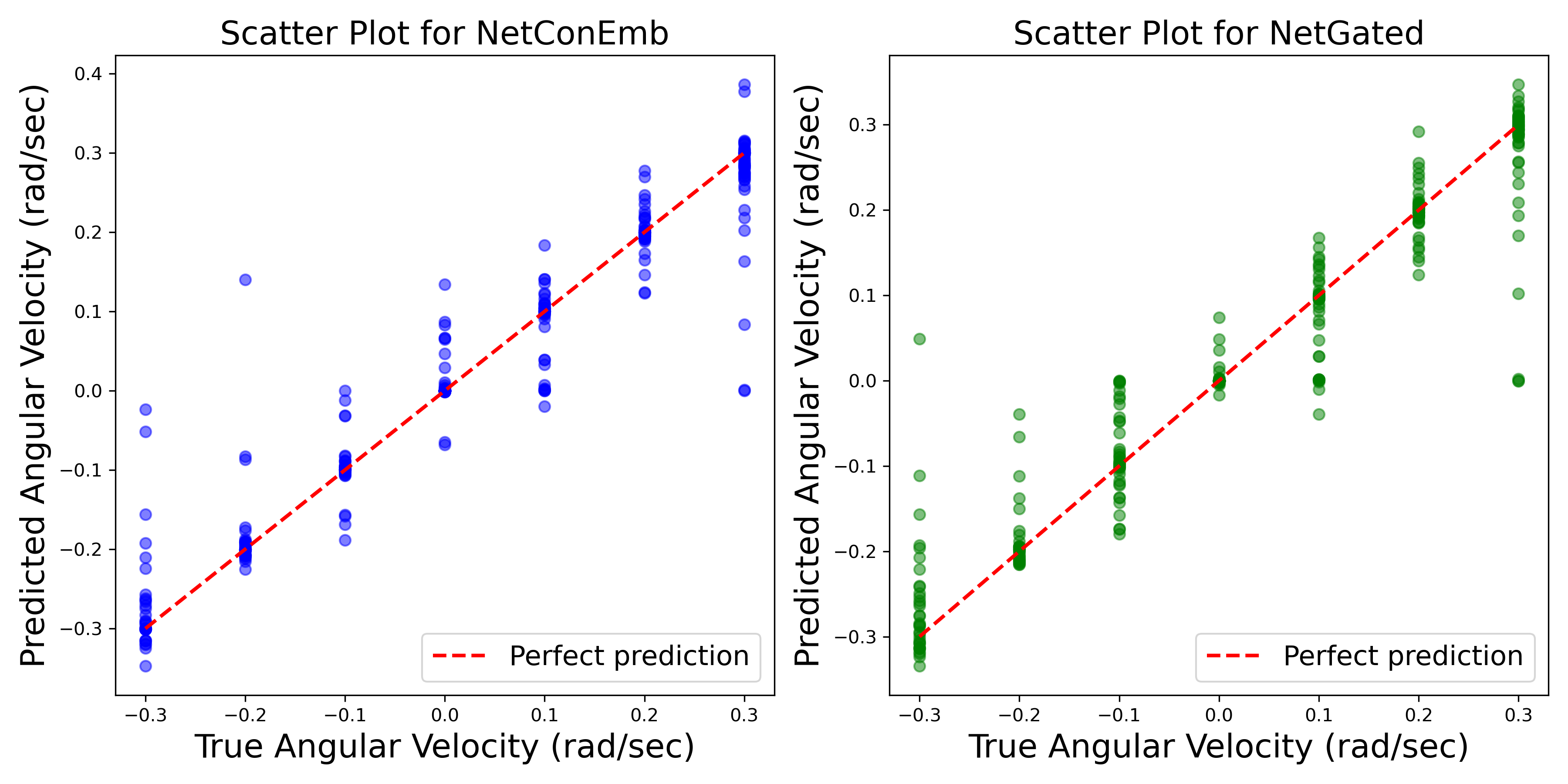}
    \caption{Scatter plot of true vs predicted angular velocity for NetConEmb and NetGated test sets.}
    \label{fig8}
\end{figure}

In addition to the quantitative results, Fig.~\ref{fig8} provides a scatter plot that shows how closely the predicted outputs align with the true values. Ideally, the points should lie along the red dashed line (which represents perfect predictions). If the points are scattered far from the line, it indicates larger prediction errors. 

\begin{figure}[t]
  \centering
    \includegraphics[width=\linewidth]{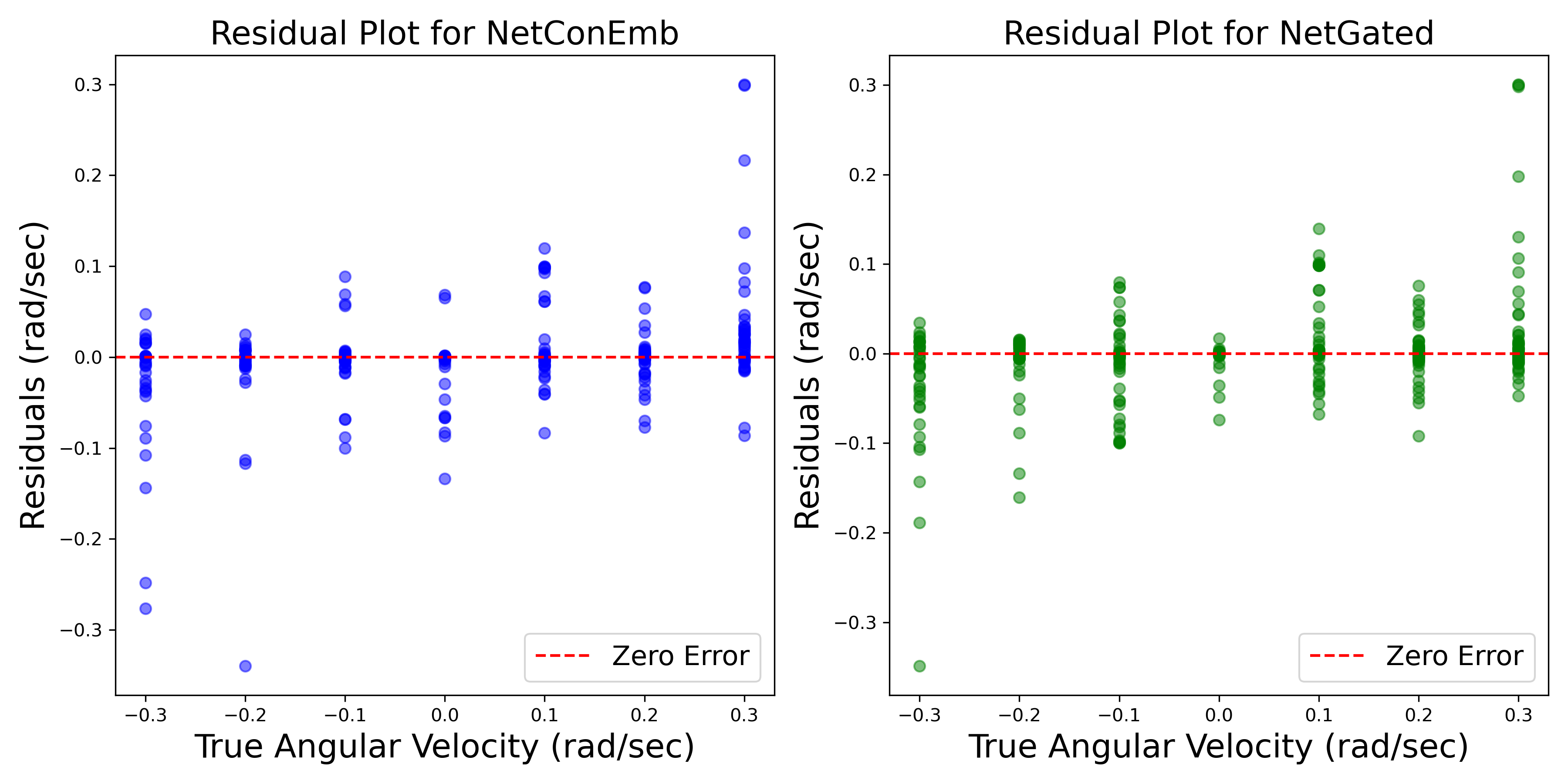}
    \caption{Residual plot for NetConEmb and NetGated test sets.}
    \label{fig9}
\end{figure}

In Fig.~\ref{fig9}, the residual plot shows the differences between the true and predicted values (residuals) against the true values. It's noted that both models generally produce residuals centered around zero, which is good. It indicates that there’s no systematic bias in the predictions (i.e., the models are not consistently under- or over-predicting). However, as the true angular velocity increases or decreases from zero, the spread of residuals increases, especially for NetGated. This indicates decreased prediction accuracy in higher velocity scenarios, which could stem from the fact that fewer training examples were used in those ranges.

Comparing both, NetGated residuals appear slightly more dispersed than NetConEmb's. This might suggest that while NetGated attempts fusion through gating, it may be more sensitive to fusion weight variations, resulting in a higher spread of prediction errors.


Ideally, the residuals should be randomly scattered around the zero line (red dashed line). Large deviations from zero indicate larger prediction errors where the model could be improved.

Based on the aforementioned analysis, the selection of a specific architecture is dependent on the situation. If efficiency and simplicity, and quick inference is more important, NetConEmb can be used. If robust performance across variable conditions is more important, the NetGated architecture should be used.

\begin{table}[b]
\caption{ Performance Comparison of NetConEmb and NetGated Models Using Color and Depth Images as Disjoint Modalities}
\label{table3}
\begin{center}
\setlength{\tabcolsep}{6pt} 
\renewcommand{\arraystretch}{1.5} 
\begin{tabular}{|p{1.5cm}|p{1.7cm}|p{0.8cm}|p{0.8cm}|p{0.8cm}|p{0.5cm}|}
\hline
\textbf{Net/Metric} & \textbf{Test Type} & \textbf{MAE $\times 10^3$} & \textbf{RMSE $\times 10^3$} & \textbf{MedAE $\times 10^3$} & \textbf{VS} \\
\hline
\multirow{2}{*}{\textbf{NetConEmb}} & Color Images & 5.47 & 24.77 & 0.05 & 0.93 \\
                                     & Depth Images & 10.26 & 38.56 & 0.4 & 0.82 \\
\hline
\multirow{2}{*}{\textbf{NetGated}}  & Color Images & 7.58 & 26.88 & 0.75 & 0.91 \\
                                     & Depth Images & 12.25 & 45.16 & 0.18 & 0.75 \\
\hline
\end{tabular}
\end{center}
\end{table}

To gain further insight into the behavior of the two models, an ablation study was conducted to evaluate their performance using depth data versus color data as the sole input for each model. In cases where only one modality was used, the values of the other modality were set to zero, effectively simulating the absence of information from that modality. A detailed comparison is presented in Table~\ref{table3} that, alongside Table~\ref{table2}, highlights the optimal approach which is to use both modalities as inputs to the network rather than relying on just one. When only depth images or color images were used, the models exhibited higher errors and lower variance scores compared to the dual-modality input. Also, it is noteworthy that both models showed clearly lower losses when color images alone were used compared to depth images alone. Lastly, the results of this ablation study, shown in Table~\ref{table3}, indicate that the NetConEmb model consistently outperforms the NetGated model, demonstrating lower losses and higher variance scores.

\section {Conclusion}
Obstacle avoidance is crucial for mobile robot navigation and vital for performing tasks in both familiar and unfamiliar settings. This study assessed two CNNs inspired by the literature with some modifications regarding the input modalities and layers used. Using a dataset of color images and depth data from a depth camera, the two CNNs were successfully trained and evaluated. Both networks successfully predicted the robot's angular velocity with minimum errors.

Results showed that NetConEmb outperformed NetGated, with lower error rates, higher variance scores, and lower inference time, suggesting that NetConEmb handles obstacle avoidance more effectively. 

The results of the ablation study showed that fusing both color and depth modalities gave better performance compared to using either modality alone.

Future research will involve deploying both networks on a mobile robot to test their real-time performance. This will help further assess their effectiveness in various environments, contributing to the development of more reliable and efficient obstacle avoidance systems for mobile robots.


\end{document}